\documentclass[10pt,letterpaper]{article}

\usepackage[letterpaper,margin=1in]{geometry}
\usepackage[T1]{fontenc}
\usepackage[utf8]{inputenc}
\usepackage{textcomp}
\usepackage{mathptmx}
\usepackage{amsmath}
\usepackage{amssymb}
\usepackage{booktabs}
\usepackage{graphicx}
\usepackage{enumitem}
\usepackage{float}
\usepackage{capt-of}
\usepackage{xcolor}
\usepackage[numbers,sort&compress]{natbib}
\definecolor{linkblue}{rgb}{0,0.08,0.45}
\usepackage[
  colorlinks=true,
  linkcolor=linkblue,
  citecolor=linkblue,
  urlcolor=linkblue,
  pdfborder={0 0 0}
]{hyperref}

\title{Elastic Queries Reinforcement Learning: Self-Aware Policy Execution for VLA Models}

\author{%
{\normalsize Ge Wang\textsuperscript{*,1,2},
Xinyu Tan\textsuperscript{*,1},
Xiang Li\textsuperscript{*,1,2},
Man Luo\textsuperscript{1},
Chengsi Yao\textsuperscript{1}}\\
{\normalsize Shenhao Yan\textsuperscript{1},
Jiahao Yang\textsuperscript{1},
Fan Feng\textsuperscript{1},
Honghao Cai\textsuperscript{1,2},
Xiangyuan Wang\textsuperscript{3}}\\
{\normalsize Zhixin Mai,
Yiming Zhao\textsuperscript{1},
Yatong Han\textsuperscript{\textdagger,1},
and Zhen Li\textsuperscript{\textdagger,2}}\\[0.35em]
{\small \textsuperscript{1}Ising AI}\\
{\small \textsuperscript{2}CUHK-Shenzhen}\\
{\small \textsuperscript{3}PKU}
}
\date{}

\hypersetup{
  pdftitle={Elastic Queries Reinforcement Learning: Self-Aware Policy Execution for VLA Models},
  pdfauthor={Ge Wang, Xinyu Tan, Xiang Li, Man Luo, Chengsi Yao, Shenhao Yan, Jiahao Yang, Fan Feng, Honghao Cai, Xiangyuan Wang, Zhixin Mai, Yiming Zhao, Yatong Han, and Zhen Li},
  pdfsubject={Preprint},
  pdfkeywords={Robot Learning, Vision-Language-Action Models, Reinforcement Learning, Adaptive Inference}
}

\newcommand{\keywords}[1]{%
  \begin{center}
    \begin{minipage}{0.92\linewidth}
      \normalsize\textbf{Keywords:} #1
    \end{minipage}
  \end{center}
}

\renewenvironment{abstract}{%
  \begin{center}
    \begin{minipage}{0.92\linewidth}
      \normalsize
      \noindent\textbf{Abstract.}\enspace
}{%
    \end{minipage}
  \end{center}
}


\newcommand{\E}{\mathbb{E}}
\newcommand{\obs}{o}

\newcommand{\gen}{F_\theta}
\newcommand{\policy}{\pi_\phi}
\newcommand{\critic}{Q_\psi}
\newcommand{\targetcritic}{Q_{\bar{\psi}}}

\begin{document}
\maketitle
\begingroup
\renewcommand{\thefootnote}{\fnsymbol{footnote}}
% Replace the placeholder emails before submitting to arXiv.
\footnotetext[1]{Equal contribution.}
\footnotetext[2]{Corresponding authors. Contact: \texttt{rstanten@alumni.stanford.edu}.}
\endgroup

\begin{abstract}
Vision-language-action (VLA) models are powerful action generators for robot manipulation, but they are typically executed with fixed inference and replanning schedules. This rigidity ignores the uneven difficulty of robot control: contact-rich or uncertain states may need more computation and fresher feedback, while easier states can often be handled with fewer inference steps and longer open-loop execution. We propose Elastic Queries Reinforcement Learning (EQRL), a framework that makes each VLA policy query elastic. A lightweight latent-schedule adaptor jointly selects the latent input, denoising budget, and action chunk length, without fine-tuning the underlying VLA model. To make scheduling difficulty-aware, EQRL trains a critic over the joint latent-schedule action and derives a state difficulty signal from critic ensemble disagreement. This signal guides compute toward difficult states, while a learned residual allows task-driven correction. We formulate variable chunk execution as query-level macro-action RL with chunk-dependent discounting and an amortized number-of-function-evaluations (NFE) budget. Across simulation and real-robot manipulation, EQRL reduces amortized inference cost while preserving or improving task success.
\end{abstract}

\keywords{Robot Learning, Vision-Language-Action Models, Reinforcement Learning, Adaptive Inference}

%===============================================================================
\section{Introduction}

Vision-language-action (VLA) policies provide reusable robot manipulation priors by combining visual perception, language-conditioned task specification, and generative action decoding~\citep{brohan2022rt,driess2023palm,zitkovich2023rt,kim2024openvla}. Recent flow-based VLAs such as $\pi_0$ generate temporally coherent action chunks from multimodal context~\citep{black2024pi_0,lipman2022flow}. Their execution interface, however, is usually rigid: each query uses a fixed denoising budget, and the robot executes a fixed number of predicted actions before replanning~\citep{zhao2023learning,chi2025diffusion}.

This fixed interface is poorly matched to manipulation, where easy free-space or approach phases alternate with contact, alignment, recovery, and precise placement. Easy states often need fewer sampling steps and can tolerate longer open-loop execution, while difficult states benefit from more computation and fresher feedback. A fixed schedule therefore wastes computation where the policy is confident and under-serves states where the decision is actually hard.

A natural way to adapt a pretrained generative action model is to keep the model frozen and learn a lightweight policy in its latent space. This preserves the behavioral prior of the VLA model and avoids expensive full-model fine-tuning. However, latent steering alone still leaves the execution schedule outside the learning problem. The adaptor may choose which latent action trajectory to generate, but the denoising budget and action-chunk length remain manually chosen hyperparameters. As a result, the learned policy cannot decide whether a given state deserves more computation, less computation, shorter commitment, or longer amortization.

\begin{figure}[tbp]
    \centering
    \hspace*{-0.04\linewidth}
    \includegraphics[
        width=0.95\linewidth,
        height=0.40\textheight,
        keepaspectratio
    ]{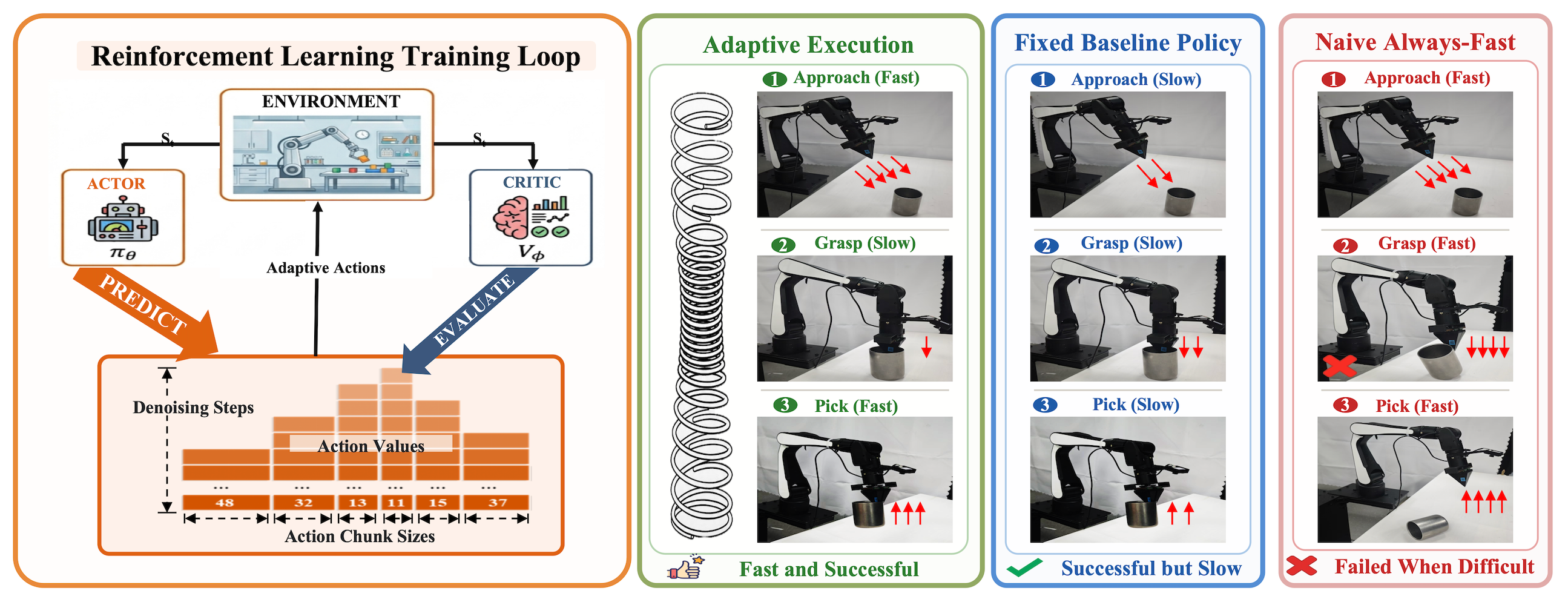}
    \caption{
    \textbf{Motivation for elastic querying.} EQRL adapts both denoising steps and execution chunk length across a task, moving quickly through easy phases while spending more feedback and compute near difficult contact phases.
    }
    \label{fig:intro_teaser}
\end{figure}

Latent-space steering can adapt a frozen generative action model without expensive full-model fine-tuning, but it leaves the execution schedule outside the learning problem. The adaptor may choose which latent trajectory to generate, while the denoising budget and action-chunk length remain manual hyperparameters. The policy therefore cannot decide whether a state deserves more computation, shorter commitment, or longer amortization.

EQRL makes each VLA query elastic. At every query, a lightweight RL adaptor selects a joint latent-schedule action $(w,K,C)$: $w$ steers the frozen sampler, $K$ controls denoising effort, and $C$ controls how many generated actions are executed before replanning. The relevant cost is therefore amortized NFE, approximately $K/C$, so the policy can reduce per-step computation in easy states while spending more feedback and compute in difficult ones.

Learning this schedule is challenging because compute penalties can dominate before the task is learned, and unconstrained schedules can collapse to constants or noisy value estimates. EQRL trains a critic over the joint latent-schedule action and derives a state-difficulty signal from critic-ensemble disagreement. This signal defines a schedule prior that allocates more denoising and shorter chunks to difficult states, while a learned residual provides task-specific corrections.

Because variable chunk lengths change the temporal scale of decisions, EQRL trains at the query level: executing a chunk induces a macro-transition, and Bellman backups use chunk-dependent discounting. An episode-level amortized NFE band controls total inference cost, while local difficulty alignment decides where the available compute should be spent.

We evaluate EQRL on LIBERO and ALOHA simulation tasks, as well as offline and online real-robot manipulation. Our contributions are threefold: (1) we formulate frozen VLA execution as query-level RL over latent steering, denoising budget, and chunk length; (2) we introduce critic-derived difficulty guidance for state-dependent compute allocation; and (3) we train variable-length chunks with macro-action Bellman backups and an episode-level amortized NFE objective. Across settings, EQRL preserves or improves success while reducing amortized inference cost, with ablations isolating the role of joint difficulty-aware scheduling.

%===============================================================================

\section{Related Work}
\paragraph{VLA models and post-training.}
Vision-language-action (VLA) models couple visual perception, language-conditioned task specification, and action generation within a unified robot policy~\citep{brohan2022rt,driess2023palm,zitkovich2023rt}.
Their progress has been driven by large-scale robot datasets, cross-embodiment training, and web-scale vision-language pretraining, which provide reusable behavioral priors across tasks and embodiments~\citep{o2024open,team2024octo,kim2024openvla,black2024pi_0}.
However, most VLA policies are still trained primarily by imitation on offline demonstrations, limiting their ability to improve through interaction under deployment-time distribution shifts~\citep{zitkovich2023rt,kim2024openvla,yang2023foundation}.
This motivates lightweight post-training methods that adapt pretrained VLA policies with task-level feedback while preserving their learned priors.

\paragraph{RL for generative robot policies.}
Reinforcement learning has recently been explored for VLA post-training, including online RL, residual adaptation, lightweight modules, and interactive post-training~\citep{lu2025vla,li2025simplevla,guo2025improving,liu2026can,tan2025interactive,xiao2025self}.
In parallel, many VLA and visuomotor policies use diffusion or flow-based action generators to model multimodal, temporally coherent behaviors~\citep{black2024pi_0,chi2025diffusion,janner2022planning}.
This connects VLA adaptation to generative-policy RL, where pretrained action generators are steered toward higher-reward behaviors without full model fine-tuning~\citep{xiao2025self,wagenmaker2025steering}.
Existing methods mainly adapt what behavior is generated, while the inference schedule used to generate and execute that behavior is usually fixed.
We build on the idea of latent-space steering, but study elastic execution for frozen flow-matching VLA action models rather than fine-tuning the generator itself.

\paragraph{Efficient and elastic execution.}
Efficient execution has been studied from the sampling side, by reducing neural function evaluations or accelerating generative action decoding~\citep{prasad2024consistency,yu2025d3p,clemente2026two,song2020denoising,lu2022dpm,song2023consistency,liu2022flow,lipman2022flow}.
A complementary line of work amortizes policy queries through action chunking, asynchronous inference, or lightweight correction of stale chunks~\citep{zhao2023learning,sendai2025leave,tang2025vlash}.
These approaches typically treat sampling effort and chunk execution as separate design choices.
In contrast, we formulate generative VLA execution as a joint elastic decision problem, where a lightweight RL adaptor selects the latent steering variable, sampling effort, and action-chunk length for each query based on the current state.

%===============================================================================

\section{Method}
\label{sec:method}

\subsection{Preliminaries}

\paragraph{Markov decision processes.}
We model robot manipulation as a Markov decision process
$\mathcal{M}=(\mathcal{S},\mathcal{A},P,p_0,r,\gamma)$.
Here $\mathcal{S}$ is the state space, $\mathcal{A}$ is the low-level action space, $P$ is the transition kernel, $p_0$ is the initial-state distribution, $r$ is the reward function, and $\gamma \in [0,1]$ is the discount factor.
For a policy $\pi$, the action-value function is
\[
    Q^\pi(s,a)
    =
    \E_\pi
    \left[
        \sum_{t=0}^{\infty}\gamma^t r(s_t,a_t)
        \mid s_0=s, a_0=a
    \right].
\]

\paragraph{Flow-based VLA models.}
Modern VLA policies often use a large vision-language backbone together with a generative action head.
Given a task-conditioned observation context $x$, the action head maps a latent/noise variable through an iterative flow or denoising process to produce an action chunk.
We denote the frozen sampling interface abstractly by
\[
    A = \Pi_\theta^{K}(x;w),
    \qquad
    A=(a^0,\ldots,a^{H-1}) \in \mathcal{A}^{H},
\]
where $w \in \mathcal{W}$ is a controllable latent/noise input exposed to the sampler and $K$ is the number of sampling steps.
This notation abstracts over the internal injection point of $w$ and over the particular visual, proprioceptive, and language conditioning used by the VLA model.
In the rest of the method, we write the current query context as $\obs$ and use $\gen$ as shorthand for this frozen sampling interface.

\paragraph{Latent-space steering.}
Latent-space steering treats the frozen generative policy as an action-space transformation: an outer policy selects a latent variable $w$, and the frozen sampler converts it into an executable action or action chunk.
This allows reinforcement learning to adapt behavior without updating the parameters of the pretrained action model.
In our setting, the learned adaptor operates on the latent-schedule interface of the frozen VLA sampler rather than on the VLA weights.

\paragraph{Problem setting.}
We study reinforcement learning adaptation of a frozen VLA sampler under repeated policy queries.
At each query, the adaptor may choose not only the latent steering variable $w$, but also the sampling budget $K$ and the number of generated actions $C$ to execute before replanning.
The resulting decision controls what action chunk is generated, how much computation is spent producing it, and how long the robot commits to it before receiving fresh feedback.

\begin{figure}[tbp]
    \centering
    \includegraphics[width=\linewidth]{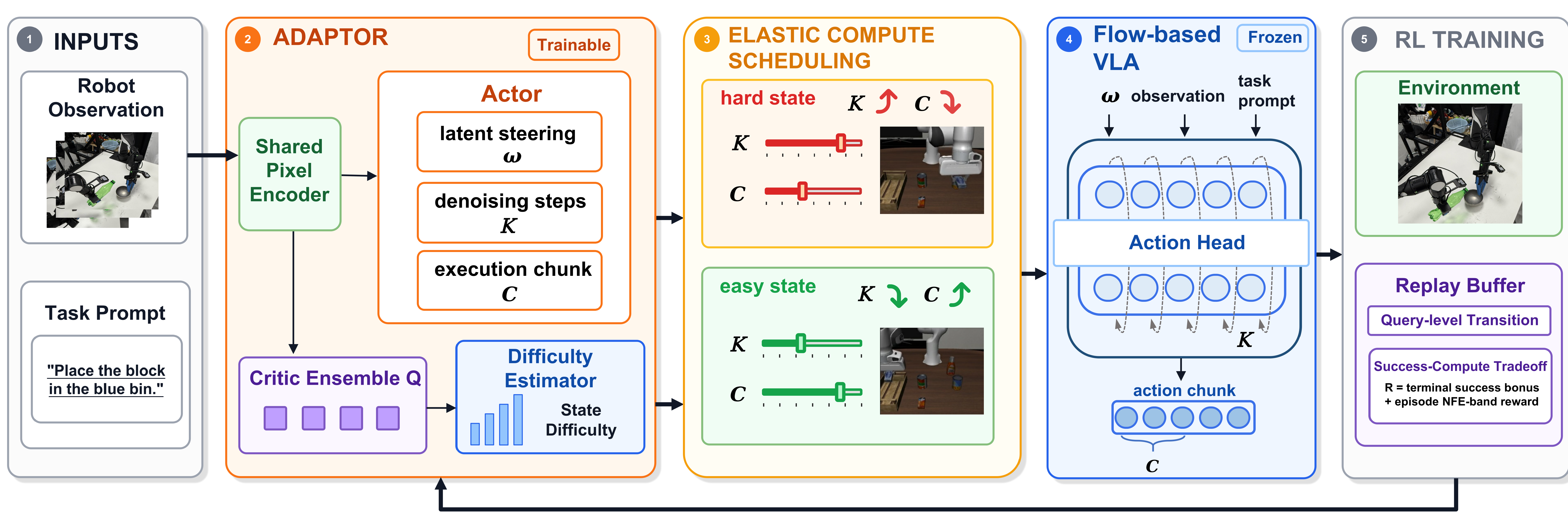}
    \caption{\textbf{EQRL architecture.}
Given the current observation and task prompt, the adaptor predicts latent steering $\omega$, denoising steps $K$, and execution chunk length $C$. Critic-derived difficulty guides the schedule toward more compute and shorter commitment in hard states, while the frozen VLA generates the executed action chunk.}
    \label{fig:method_overview}
\end{figure}

\subsection{Latent-Schedule Query Actions}

Let $j$ index policy queries, and let $t_j$ be the environment time at which query $j$ begins.
The adaptor has a shared observation encoder followed by three stochastic heads, and samples
\[
    u_j = (w_j,z^K_j,z^C_j) \sim \policy(\cdot \mid \obs_{t_j}),
\]
where $w_j$ is the latent steering variable and $z^K_j,z^C_j \in [-1,1]$ are continuous schedule variables.
The schedule variables do not directly define the executed sampling budget and chunk length.
Instead, they define learned residual proposals around the base schedule.
We map $z^K_j$ and $z^C_j$ to bounded residuals $\Delta K_j$ and $\Delta C_j$ around the base schedule $(K_0,C_0)$, with ranges constrained by $[K_{\min},K_{\max}]$ and $[C_{\min},C_{\max}]$.
The next subsection combines these learned residuals with a critic-derived schedule prior to obtain the final schedule.

\subsection{Difficulty-Guided Schedule Parameterization}

The critic is an ensemble of state-action value functions
\[
    \{\critic^{(i)}(\obs,\tilde u)\}_{i=1}^{M},
\]
defined over the joint latent-schedule action.
Here $\tilde u=[w,\bar K,\bar C]$ denotes a latent variable together with normalized continuous schedule coordinates, and $\bar K_0,\bar C_0$ denote the normalized base schedule.
It evaluates a latent steering variable together with the sampling budget and temporal commitment used to execute the generated chunk.
To make scheduling state-aware, we derive a difficulty signal from critic ensemble disagreement at the base schedule.
Let
\[
    w_\pi(\obs) =
    \operatorname{mode}\left(\pi_w(\cdot\mid\obs)\right).
\]
We compute
\[
    q_{\mathrm{std}}(\obs) =
    \operatorname{Std}_{i=1}^{M}
    \left[
        \critic^{(i)}\left(
        \obs,
        [w_\pi(\obs),\bar K_0,\bar C_0]
        \right)
    \right].
\]
The raw disagreement score is normalized before it controls the schedule.
During rollout we use running statistics, and during updates we use batch statistics of the same score.
Denote the normalized score by $d(\obs)$.
We convert it to a centered rank signal
\[
    r_c(\obs)
    =
    2\,\operatorname{rank}(d(\obs))-1
    \in [-1,1],
\]

where $\operatorname{rank}(\cdot)$ denotes a percentile rank in $[0,1]$.
The signed difficulty is
\[
    s(\obs)
    =
    \tanh\left(\tau\,r_c(\obs)\right)
    \in [-1,1].
\]
Positive values indicate states that should receive more conservative execution, while negative values indicate states that can be handled with lower amortized computation.

The signed difficulty defines a schedule prior.
Let $[x]_+ = \max(x,0)$.
The sampling-step prior is
\[
K^{p}(\obs) =
K_0
+ [s(\obs)]_+(K_{\max}-K_0)
- [-s(\obs)]_+(K_0-K_{\min}),
\]
and the chunk prior is
\[
C^{p}(\obs) =
C_0
- [s(\obs)]_+(C_0-C_{\min})
+ [-s(\obs)]_+(C_{\max}-C_0).
\]
Thus high-difficulty states receive more sampling steps and shorter chunks, while low-difficulty states receive fewer sampling steps and longer chunks.

The final continuous schedule is the sum of this prior and the learned residual:
\[
    \hat K_j =
    K^{p}(\obs_{t_j})
    + \lambda_r g_j \Delta K_j,
    \qquad
    \hat C_j =
    C^{p}(\obs_{t_j})
    + \lambda_r g_j \Delta C_j,
\]
where $g_j \in [0,1]$ is a schedule gate and $\lambda_r$ controls residual authority.
The continuous values are clipped to their valid ranges and rounded for environment interaction, yielding integer execution values $K_j$ and $C_j$.
The frozen sampler is then queried with $(\obs_{t_j},w_j,K_j)$, and the robot executes the first $C_j$ generated actions unless the episode terminates earlier.
The critic uses the final continuous schedule targets
\[
    \tilde u_j = [w_j,\bar K_j,\bar C_j],
\]
\[
    \bar K_j =
    2\frac{\hat K_j-K_{\min}}{K_{\max}-K_{\min}}-1,
    \qquad
    \bar C_j =
    2\frac{\hat C_j-C_{\min}}{C_{\max}-C_{\min}}-1.
\]
Using continuous targets in the critic preserves a smooth training signal, while integer rounding is confined to environment interaction.

\subsection{Query-Level Learning Objective}
\label{sec:training_budgeting}

Executing a chunk induces a query-level macro transition.
Let $L_j \leq C_j$ denote the actual number of executed low-level actions.
The replay buffer stores
\[
    \left(\obs_{t_j},\tilde u_j,R_j,\obs_{t_j+L_j},L_j,d_j\right),
\]
where $d_j$ is the terminal flag.
Because one query can cover multiple environment steps, Bellman backups use the chunk-dependent discount $\gamma^{L_j}$.

For sparse-success manipulation tasks, we assign task reward at the query level by giving a terminal success bonus to the final query of a successful episode.
In addition, an episode-level compute-budget reward is distributed uniformly over all queries in the episode:
\[
    R_j =
    b_{\mathrm{ep}}
    +
    \mathbf{1}[j=N_{\mathrm{ep}}]\mathbf{1}[\mathrm{success}],
    \qquad
    b_{\mathrm{ep}}
    =
    -\lambda_{\mathrm{cost}}
    \frac{P_{\mathrm{ep}}}{N_{\mathrm{ep}}}.
\]
We use the sign convention that positive $P_{\mathrm{ep}}$ is a compute cost and negative $P_{\mathrm{ep}}$ is a saving bonus.

For a next action $\tilde u' \sim \policy(\cdot \mid \obs')$, the Bellman target is
\[
    y =
    R
    +
    \gamma^{L}(1-d)
    \bar Q_{\mathrm{targ}}(\obs',\tilde u'),
    \qquad
    \bar Q_{\mathrm{targ}}(\obs,\tilde u)
    =
    \frac{1}{M}\sum_{i=1}^{M} \targetcritic^{(i)}(\obs,\tilde u).
\]
We fit the critic ensemble by minimizing a mean-squared Bellman error. The actor is optimized with the maximum-entropy latent-schedule objective~\citep{haarnoja2018soft}
\[
    \mathcal{L}_{\mathrm{sac}}
    =
    \E_{\obs,u \sim \policy}
    \left[
        \alpha \log \policy(u\mid \obs)
        -
        \bar Q(\obs,\tilde u)
    \right],
    \qquad
    \bar Q(\obs,\tilde u) =
    \frac{1}{M}\sum_{i=1}^{M}\critic^{(i)}(\obs,\tilde u).
\]

We also regularize local compute allocation to follow the critic-derived difficulty.
Let
\[
    a_{\mathrm{alloc}} =
    \bar K^{01}-\bar C^{01},
\]
where $\bar K^{01}$ and $\bar C^{01}$ are normalized to $[0,1]$ over the current schedule range.
Large $a_{\mathrm{alloc}}$ corresponds to more sampling steps and shorter commitment; small $a_{\mathrm{alloc}}$ corresponds to fewer sampling steps and longer commitment.
An auxiliary alignment loss encourages $a_{\mathrm{alloc}}$ to have the same sign as the difficulty signal $s$, so that difficult states receive higher compute and shorter chunks while easy states receive lower amortized compute.

Global compute pressure is imposed through an episode-level NFE band.
For an episode with query budgets $\{K_j,C_j\}_{j=1}^{N_{\mathrm{ep}}}$, the requested amortized NFE is
\[
    \bar\rho_{\mathrm{ep}}
    =
    \frac{\sum_{j=1}^{N_{\mathrm{ep}}} K_j}
         {\sum_{j=1}^{N_{\mathrm{ep}}} C_j}.
\]
Episodes above the upper band incur an over-budget cost, with a small tolerance for budget debt.
Successful episodes may also receive an under-use cost if computation falls below the lower band, and a small negative cost when they remain inside the target band.
These terms define $P_{\mathrm{ep}}$, which is distributed uniformly over query transitions through $b_{\mathrm{ep}}$ in the reward definition above.
This separates the two roles of scheduling: the NFE band controls the total amount of computation, while the difficulty-alignment term determines where that computation should be spent.

%===============================================================================

\section{Experiments}
\label{sec:experiments}

Our experiments address three empirical questions: 
(1) whether difficulty-aware elastic querying can reduce inference cost while preserving, or even improving, task success in simulation;
(2) whether EQRL is suitable for real-robot learning under physical execution constraints; and
(3) whether the gains come from actor--critic difficulty awareness and joint scheduling, rather than simply lowering the compute budget or adapting only one schedule dimension.

\begin{table}[tbp]
    \centering
    \caption{Simulation success and efficiency on LIBERO-4Tasks and ALOHA-Cube. EQRL improves the best baseline success/AUC on both benchmarks while reducing relative NFE by 24.3\% on LIBERO and 20.0\% on ALOHA.}
    \label{tab:sim_efficiency}
    \begin{tabular*}{\textwidth}{@{\extracolsep{\fill}}llccc}
        \toprule
        Env. & Method 
        & Final Success $\uparrow$ 
        & Success AUC $\uparrow$ 
        & Rel. NFE $\downarrow$ \\
        \midrule
        LIBERO-4Tasks 
        & DSRL  
        & 0.95 
        & 0.86 
        & 0.500 \\
        
        & NOISE 
        & 0.93 
        & 0.68 
        & 0.400 \\
        
        & SDE   
        & 0.83 
        & 0.69 
        & 0.400 \\
        
        & \textbf{EQRL} 
        & \textbf{0.96} {\scriptsize(+0.01)}
        & \textbf{0.90} {\scriptsize(+0.04)}
        & \textbf{0.303} {\scriptsize(24.3\%$\downarrow$)} \\
        
        \midrule
        ALOHA-Cube 
        & DSRL  
        & 0.94 
        & 0.81 
        & 0.200 \\
        
        & NOISE 
        & 0.91 
        & 0.71 
        & 0.200 \\
        
        & SDE 
        & 0.85 
        & 0.73 
        & 0.200 \\
        
        & \textbf{EQRL} 
        & \textbf{1.00} {\scriptsize(+0.06)}
        & \textbf{0.82} {\scriptsize(+0.01)}
        & \textbf{0.160} {\scriptsize(20.0\%$\downarrow$)} \\
        
        \bottomrule
    \end{tabular*}
\end{table}

\subsection{Experimental Setup}

\paragraph{Settings.}
We evaluate on LIBERO and ALOHA simulation tasks, as well as offline and online real-robot manipulation~\citep{liu2023libero,zhao2023learning}.
All methods use the same $\pi_0$ VLA policy~\citep{black2024pi_0} as the base action model. Additional implementation details, task descriptions, hyperparameters, and evaluation protocols are provided in Appendix~\ref{app:experimental_details}.

\paragraph{Baselines.}
We compare behavior cloning (BC/SFT), DSRL~\citep{wagenmaker2025steering}, RLinf-Noise and RLinf-SDE, and EQRL.
RLinf-Noise and RLinf-SDE are the RLinf implementations~\citep{yu2025rlinf} of Flow-GRPO~\citep{liu2026flow} and ReinFlow~\citep{zhang2026reinflow}, respectively.
EQRL jointly selects the latent input, denoising budget, and action chunk length at each query.

\paragraph{Metrics.}
We report task success, success AUC, and the number of environment steps required to reach high-success thresholds such as 80\% or 90\%.
For efficiency, we report average denoising steps, average requested chunk length, amortized NFE, NFE reduction, and real-robot wall-clock latency.
For schedule diagnostics, we compare selected denoising steps and chunk lengths across critic-derived difficulty levels.

\subsection{Difficulty-Aware Acceleration in Simulation}

\begin{figure}[t]
    \centering
    \begin{minipage}[t]{0.249\linewidth}
        \centering
        \includegraphics[width=\linewidth]{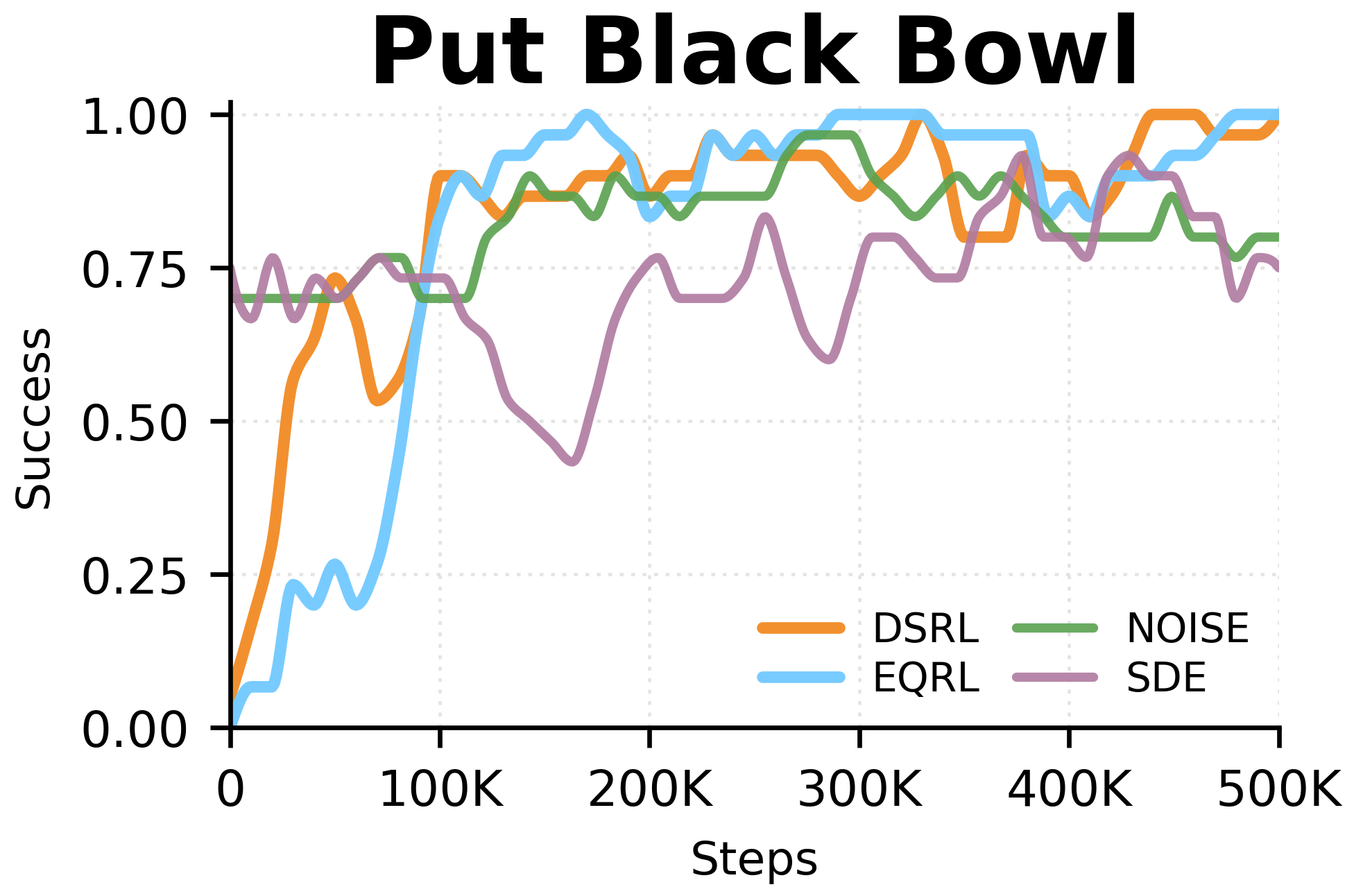}
    \end{minipage}%
    \hspace{0pt}%
    \begin{minipage}[t]{0.249\linewidth}
        \centering
        \includegraphics[width=\linewidth]{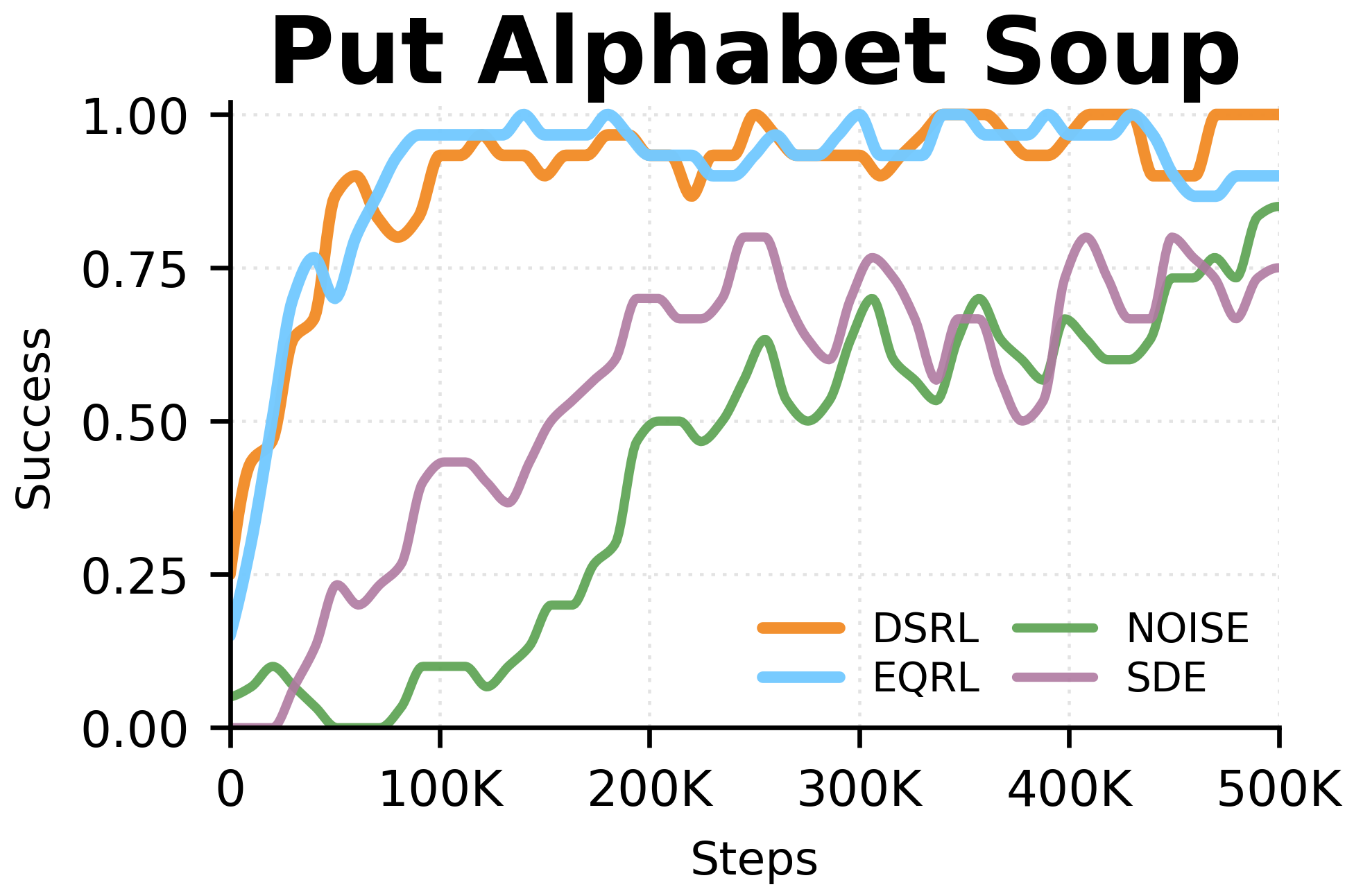}
    \end{minipage}%
    \hspace{0pt}%
    \begin{minipage}[t]{0.249\linewidth}
        \centering
        \includegraphics[width=\linewidth]{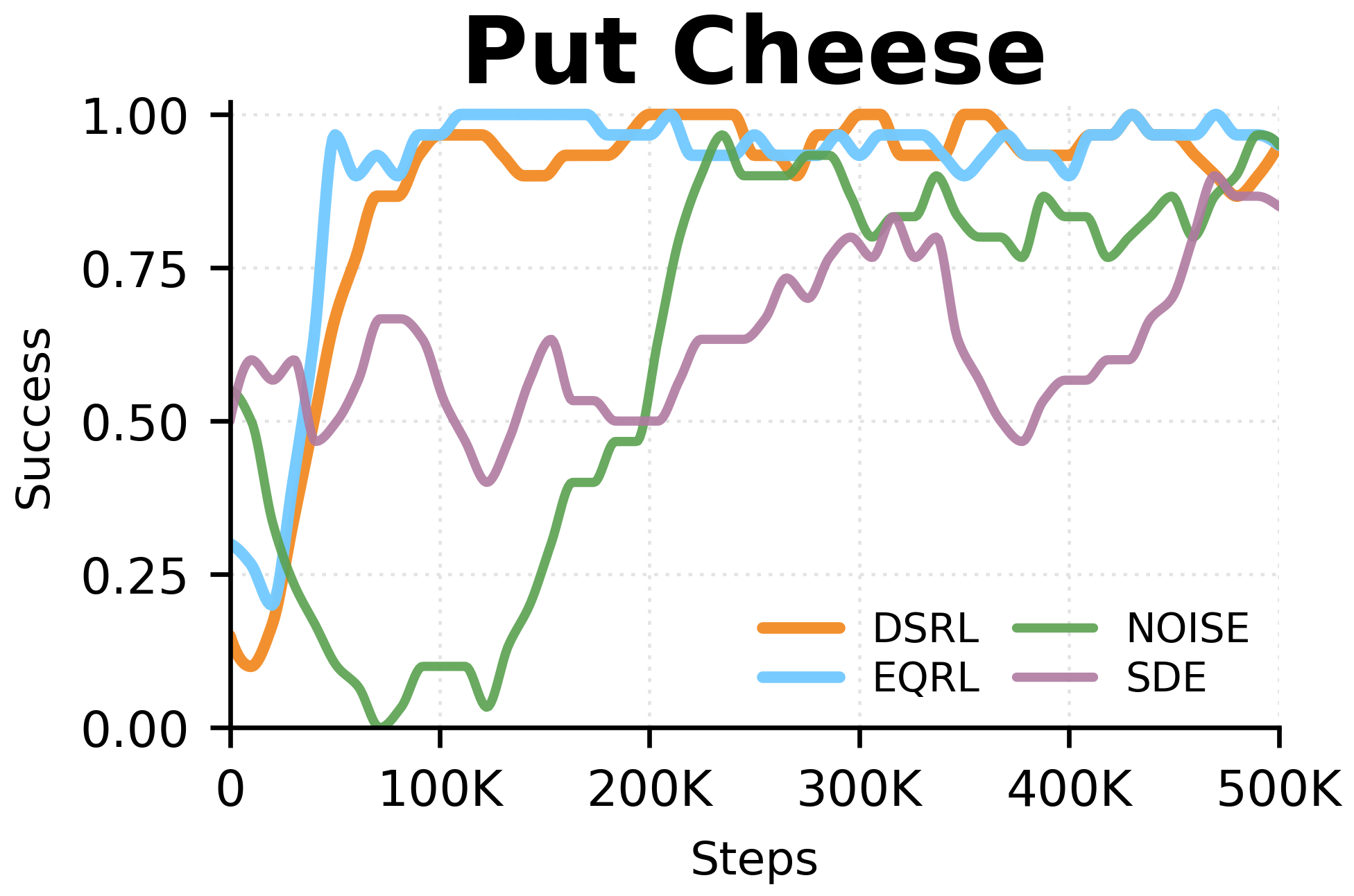}
    \end{minipage}%
    \hspace{0pt}%
    \begin{minipage}[t]{0.249\linewidth}
        \centering
        \includegraphics[width=\linewidth]{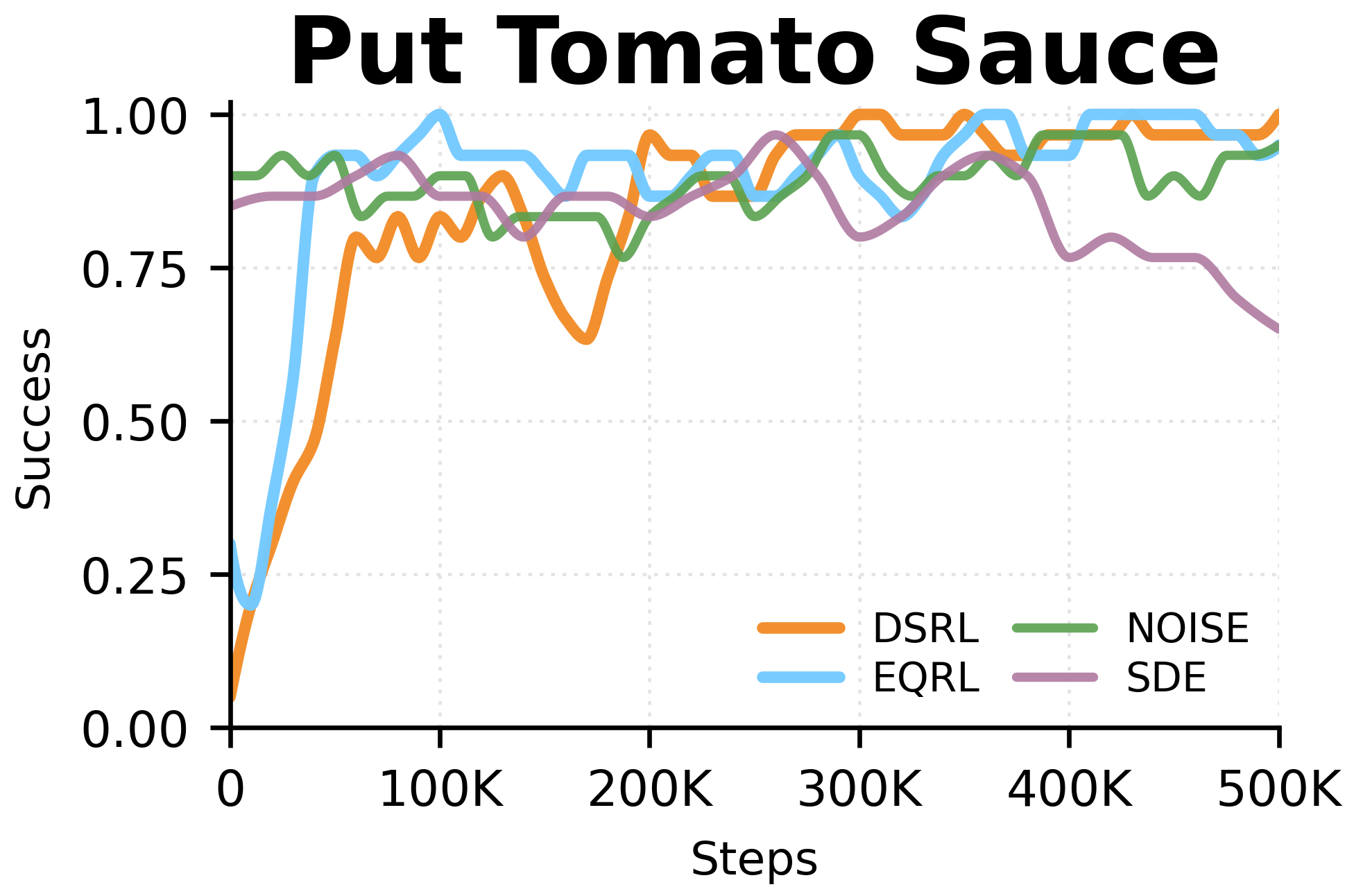}
    \end{minipage}

    \caption{LIBERO success over 500k training steps on LIBERO-90. EQRL reaches high success quickly on several tasks and improves aggregate success AUC while reducing relative NFE.}
    \label{fig:libero_success}
\end{figure}

\noindent
\begin{minipage}[t]{0.53\linewidth}
    \vspace{0pt}
    We first test whether difficulty-aware elastic querying improves the success--cost trade-off in simulation.

    Figure~\ref{fig:libero_success} and Figure~\ref{fig:aloha_success} show the learning behavior on LIBERO and ALOHA, while Table~\ref{tab:sim_efficiency} summarizes final success, success AUC, and relative NFE.
    Across both settings, EQRL preserves or improves task success while reducing inference cost.
    The success AUC improvement on LIBERO suggests that elastic scheduling helps the learning process, rather than merely lowering the final execution budget.

    The ALOHA result further indicates that this benefit is not specific to one benchmark family.
    The ablation study later tests whether this comes from critic-aware allocation, rather than simply reducing the average compute budget.
\end{minipage}\hfill
\begin{minipage}[t]{0.42\linewidth}
    \vspace{0pt}
    \centering
    \includegraphics[width=\linewidth]{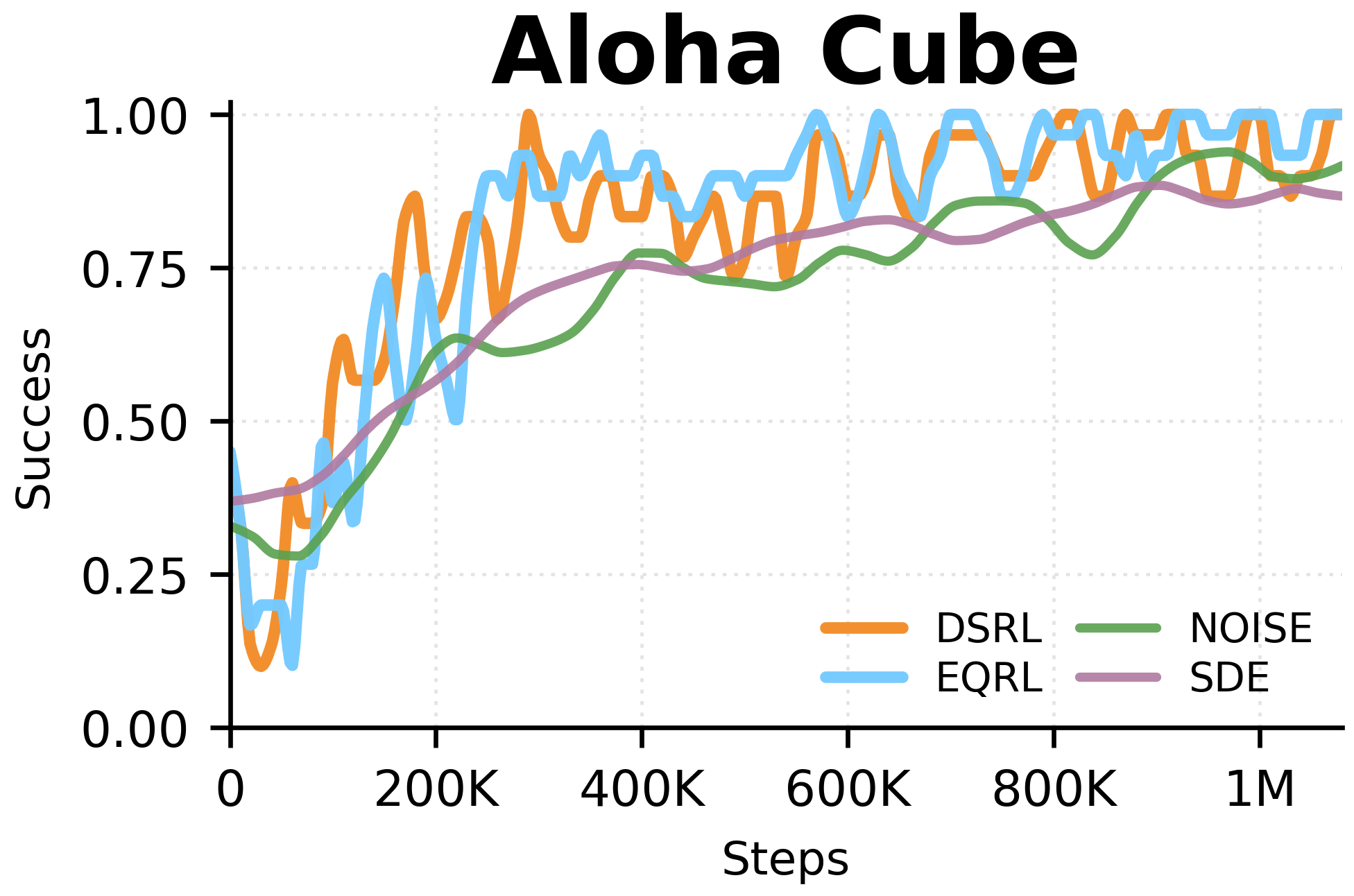}
    \captionof{figure}{ALOHA Cube success over training. EQRL reaches perfect final success while using lower relative NFE than fixed-schedule baselines.}
    \label{fig:aloha_success}
\end{minipage}

\subsection{Real-world Reinforcement Learning with EQRL}
\noindent
\begin{minipage}[t]{0.50\linewidth}
    \vspace{0pt}
    We next evaluate whether EQRL remains suitable for real-robot learning, where scheduling decisions affect not only NFE but also wall-clock latency and physical execution stability.
    For offline adaptation, we train each task independently for 500k gradient steps using 50--100 expert trajectories collected on DISCOVER Robotics PTK2 platforms.

    For online RL, we use a DISCOVER Robotics TOK4 platform, initialize from 20 expert trajectories, and stop after approximately 15k real-environment interaction steps.

    Figure~\ref{fig:real_robot} visualizes how EQRL allocates computation during a real-world pouring trajectory.
    Table~\ref{tab:real_robot_all}(a) shows that EQRL reduces relative NFE and trajectory time while maintaining comparable average task success across offline real-robot tasks, and Table~\ref{tab:real_robot_all}(b) shows that the same interface remains effective when the adaptor is updated from real interaction.
    Together, these results indicate that elastic querying can translate computational savings into practical speedups and improve the contact-time trade-off by learning when to commit and when to replan.
\end{minipage}\hfill
\begin{minipage}[t]{0.46\linewidth}
    \vspace{0pt}
    \centering
    \includegraphics[
        width=\linewidth,
        height=0.21\textheight,
        keepaspectratio
    ]{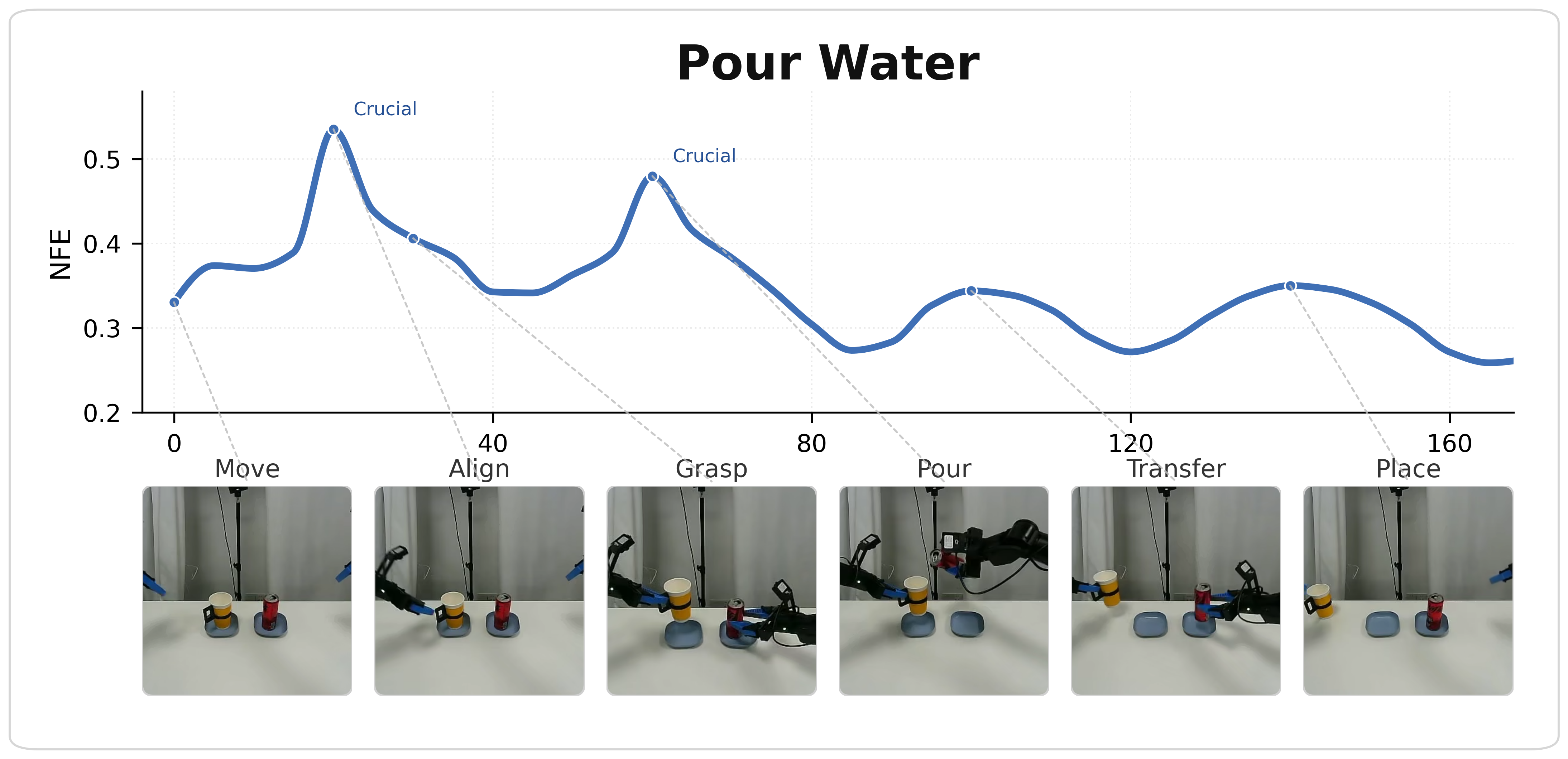}
    \captionof{figure}{Real-robot elastic execution and learning results. The pouring timeline shows EQRL increasing NFE near crucial alignment/contact phases and lowering it during steadier motion, while the offline and online comparisons show comparable or higher success with lower relative NFE than DSRL.}
    \label{fig:real_robot}
\end{minipage}

\subsection{Ablating Actor--Critic Difficulty Awareness}

\begin{figure}[t]
    \centering
    \begin{minipage}[t]{0.235\linewidth}
        \centering
        \includegraphics[width=\linewidth]{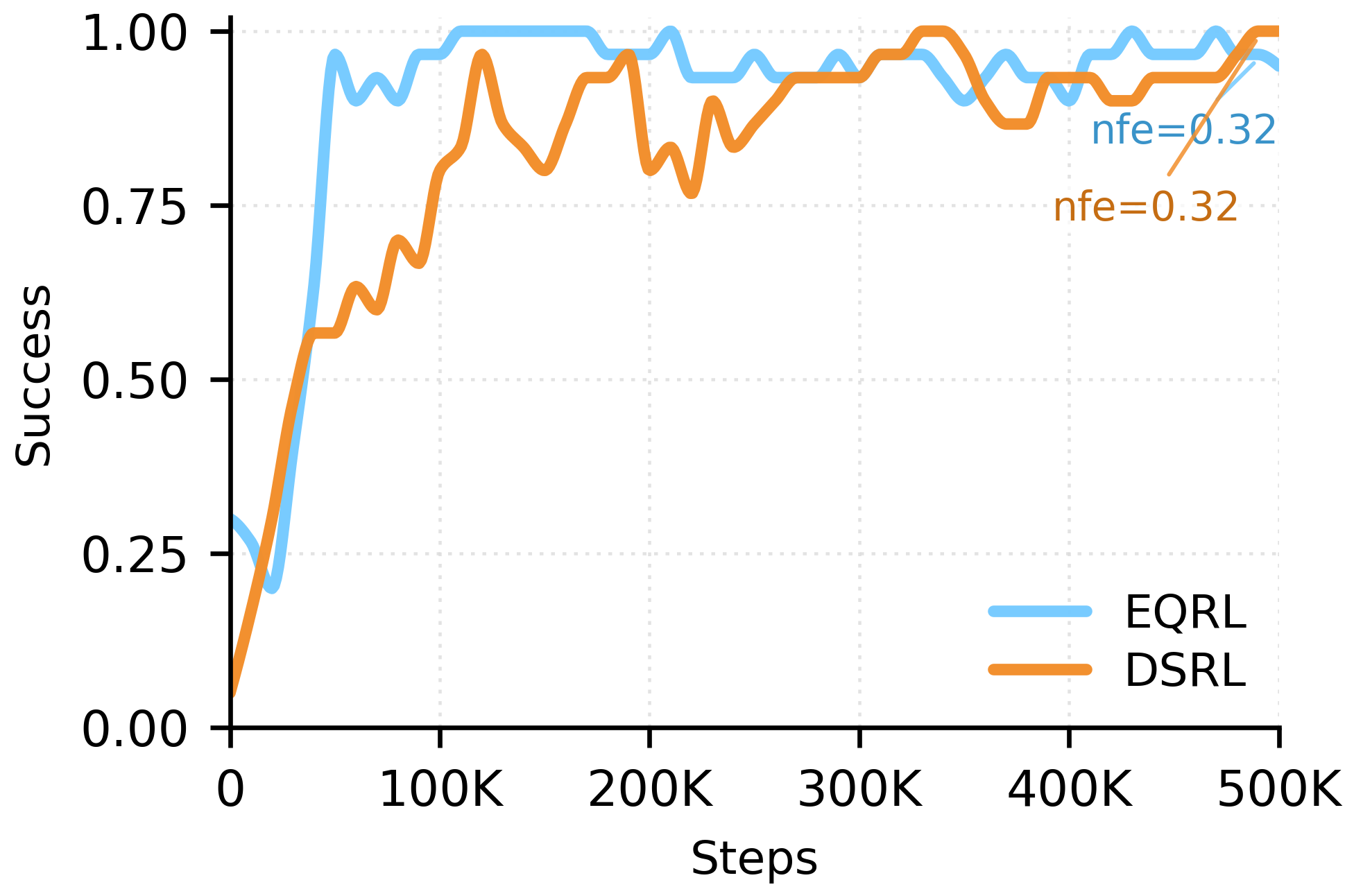}
    \end{minipage}
    \hfill
    \begin{minipage}[t]{0.235\linewidth}
        \centering
        \includegraphics[width=\linewidth]{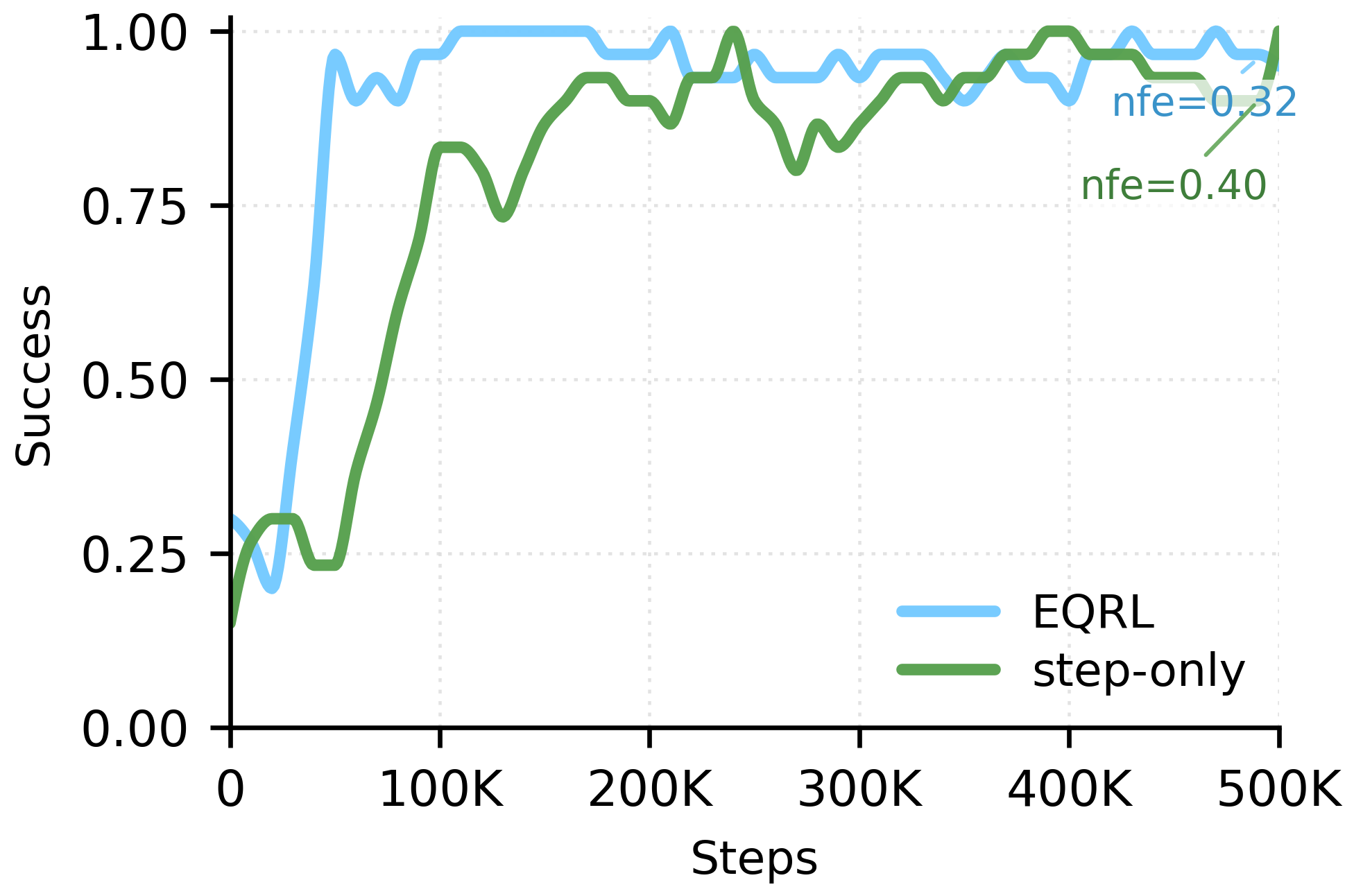}
    \end{minipage}
    \hfill
    \begin{minipage}[t]{0.235\linewidth}
        \centering
        \includegraphics[width=\linewidth]{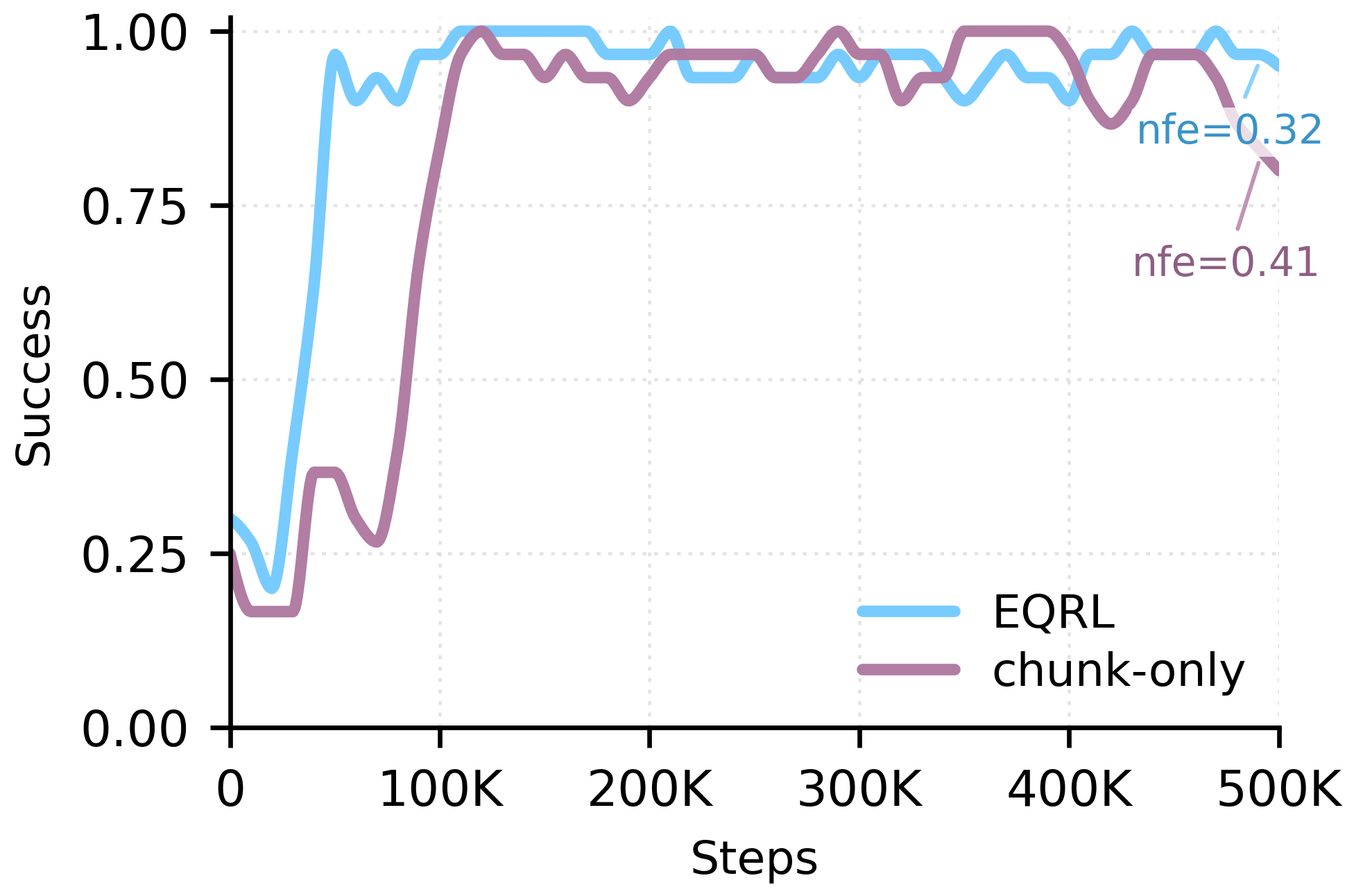}
    \end{minipage}
    \hfill
    \begin{minipage}[t]{0.235\linewidth}
        \centering
        \includegraphics[width=\linewidth]{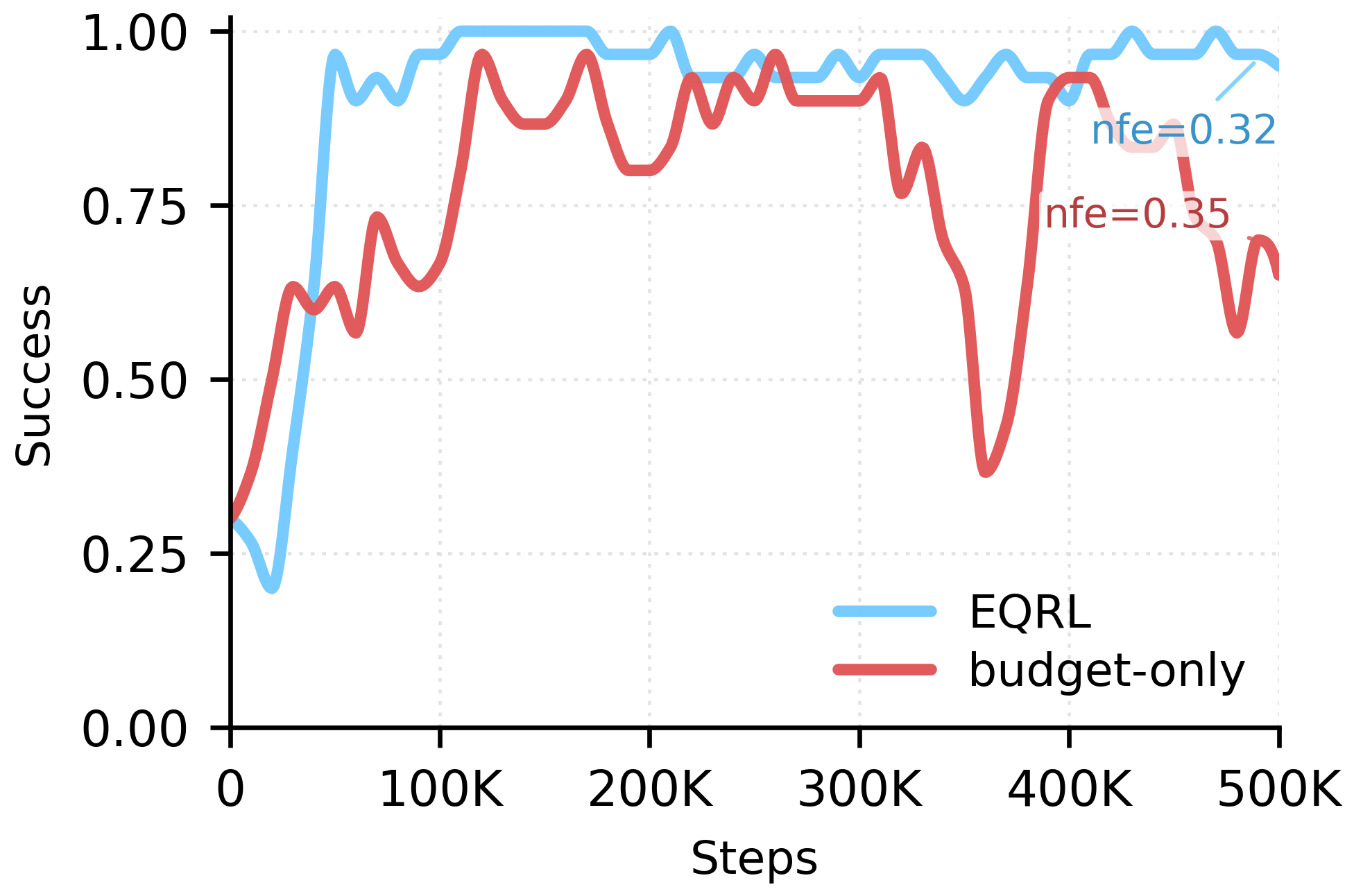}
    \end{minipage}
    \caption{Ablation study on Put Cheese. From left to right, the panels compare the full method against matched-NFE DSRL, dynamic-steps-only, dynamic-chunks-only, and budget-only variants.}
    \label{fig:ablations}
\end{figure}

We use the Put Cheese task as a diagnostic setting to test the source of EQRL's gains.
Figure~\ref{fig:ablations} compares the full method with matched-NFE DSRL, dynamic-steps-only, dynamic-chunks-only, and budget-only variants.

\noindent
\begin{minipage}[t]{0.45\linewidth}
    \vspace{0pt}
    The comparison shows that reducing average NFE is not sufficient by itself.
    Although matched-NFE DSRL and the budget-only variant operate under a similar compute target, they lag behind in their success-rate trajectories, indicating that the improvement is not simply caused by using a smaller fixed budget.
    Likewise, adapting only denoising steps or only chunk length weakens performance, suggesting that compute allocation and replanning frequency need to be coordinated.

    The ablations indicate that EQRL needs both actor--critic difficulty awareness and joint scheduling.
    The critic-aware schedule saves computation on easier phases while preserving denoising and feedback in harder phases, avoiding the success degradation seen in the restricted variants.
\end{minipage}\hfill
\begin{minipage}[t]{0.51\linewidth}
    \vspace{0pt}
    \centering
    \captionof{table}{Real-robot evaluation. Parentheses report EQRL differences relative to DSRL.}
    \label{tab:real_robot_all}

    \vspace{0.15em}
    {\footnotesize\textbf{(a) Offline comparison.}}
    \vspace{0.05em}

    \scriptsize
    \setlength{\tabcolsep}{1.0pt}
    \renewcommand{\arraystretch}{0.88}
    \begin{tabular*}{\linewidth}{@{\extracolsep{\fill}}llccc}
        \toprule
        Task & Method & Succ. $\uparrow$ & NFE $\downarrow$ & Time $\downarrow$ \\
        \midrule
        Pour & DSRL & 0.55 & 0.500 & 8.94 \\
        & \textbf{EQRL}
        & \textbf{0.75} {\tiny(+.20)}
        & \textbf{0.332} {\tiny(33.6\%$\downarrow$)}
        & \textbf{7.36} {\tiny(17.7\%$\downarrow$)} \\
        \midrule
        Fruit & DSRL & 0.85 & 0.500 & 6.91 \\
        & \textbf{EQRL}
        & \textbf{0.85} {\tiny(+.00)}
        & \textbf{0.345} {\tiny(31.0\%$\downarrow$)}
        & \textbf{5.31} {\tiny(23.2\%$\downarrow$)} \\
        \midrule
        Cucumber & DSRL & 0.95 & 0.500 & 5.98 \\
        & \textbf{EQRL}
        & \textbf{0.90} {\tiny(-.05)}
        & \textbf{0.334} {\tiny(33.2\%$\downarrow$)}
        & \textbf{5.29} {\tiny(11.5\%$\downarrow$)} \\
        \midrule
        Handover & DSRL & 0.90 & 0.500 & 6.76 \\
        & \textbf{EQRL}
        & \textbf{0.85} {\tiny(-.05)}
        & \textbf{0.341} {\tiny(31.9\%$\downarrow$)}
        & \textbf{6.02} {\tiny(11.0\%$\downarrow$)} \\
        \midrule
        Avg. & DSRL & 0.813 & 0.500 & 7.15 \\
        & \textbf{EQRL}
        & \textbf{0.838} {\tiny(+.025)}
        & \textbf{0.338} {\tiny(32.4\%$\downarrow$)}
        & \textbf{6.00} {\tiny(16.1\%$\downarrow$)} \\
        \bottomrule
    \end{tabular*}

    \vspace{0.35em}
    {\footnotesize\textbf{(b) Online comparison.}}
    \vspace{0.05em}

    \setlength{\tabcolsep}{2pt}
    \renewcommand{\arraystretch}{0.90}
    \begin{tabular*}{\linewidth}{@{\extracolsep{\fill}}lcc}
        \toprule
        Method & Success Rate $\uparrow$ & Rel. NFE $\downarrow$ \\
        \midrule
        DSRL & 0.90 & 0.500 \\
        \textbf{EQRL}
        & \textbf{0.95} {\tiny(+0.05)}
        & \textbf{0.341} {\tiny(31.8\%$\downarrow$)} \\
        \bottomrule
    \end{tabular*}
\end{minipage}

%===============================================================================
\section{Discussion and Limitations}

Our results suggest that VLA execution is a state-dependent resource-allocation problem.
EQRL makes this structure visible through the learned actor and critic: the policy allocates fewer denoising steps and longer chunks to easier states, while using more steps and shorter chunks near difficult interaction phases.
This elastic behavior reduces amortized inference time and can improve success by spending extra computation only when the state requires it; the real-robot results further suggest that elastic compute and replanning are useful execution-time interfaces for frozen VLA policies.

There are also several limitations.
First, offline training cannot directly observe the causal effect of different denoising budgets and chunk lengths, because expert datasets provide fixed trajectories without environment feedback under alternative schedules.
Although our offline setting can compare the accuracy of chunks generated with different $K$ and $C$ against expert actions, this proxy is less reliable than task reward from real environment interaction.
Second, real-robot training is still not sufficiently efficient or general.
In practice, the adaptor is fine-tuned per task, and the learned schedule does not yet provide strong task-level generalization across substantially different manipulation skills.

%===============================================================================

\clearpage

\bibliography{references}

\newpage
\section{APPENDIX}
\label{appendix}

\subsection{Experimental Details}
\label{app:experimental_details}

All methods share the same frozen $\pi_0$ base policy~\citep{black2024pi_0} and a
Pixel-SAC agent that acts in the noise latent space. Our method (E-QRL) additionally
learns an elastic schedule, i.e., the injected noise $w$ together with the number
of denoising steps and the action-chunk length, under a compute (NFE) budget and a
$q_\mathrm{std}$-based difficulty signal. The DSRL baseline~\citep{wagenmaker2025steering}
uses the same SAC backbone but keeps the denoising steps and action chunk fixed and
removes all elastic modules (range, NFE budget, and difficulty). We
report the hyperparameters for the real-robot experiments (Table~\ref{tab:hp_airbot}),
the LIBERO simulation tasks (Table~\ref{tab:hp_libero}), and the ALOHA simulation task
(Table~\ref{tab:hp_aloha}).

Hyperparameters whose meaning is shared across environments use the same row name in
all tables. ``Training steps'' denotes gradient steps for the offline setting and
environment-interaction steps for the online settings (both capped by \texttt{max\_steps}).
``Grad steps / collection'' is the number of SAC updates performed after each data
collection round (\texttt{train\_steps\_per\_collection} for the online real-robot setting,
\texttt{multi\_grad\_step} for simulation; the offline setting has no collection round).
For the DSRL baseline, ``--'' marks hyperparameters that are specific to E-QRL and have
no counterpart in the baseline.

% ---------------------------------------------------------------------------
% Table 1: Real-robot (DISCOVER Robotics PTK2 / TOK4)
% ---------------------------------------------------------------------------
\begin{table}[H]
\centering
\caption{Hyperparameters for the real-robot E-QRL experiments}
\label{tab:hp_airbot}
\small
\begin{tabular}{lll}
\toprule
Hyperparameter & Offline & Online \\
\midrule
Task                              & grasp\_fruit & Pouring Water \\
Base policy                       & $\pi_0$, frozen & $\pi_0$, frozen \\
Train mode                        & Offline & Online \\
Training steps                    & $5\times10^{5}$ & $5\times10^{5}$ \\
Batch / discount $\gamma$         & 256 / 0.999 & 256 / 0.999 \\
LR (actor / critic / temp)        & $10^{-4}$ / $3{\times}10^{-4}$ / $3{\times}10^{-4}$ & $10^{-4}$ / $3{\times}10^{-4}$ / $3{\times}10^{-4}$ \\
Critic ensemble \texttt{num\_qs}  & 10 & 10 \\
Action magnitude                  & 2.0 & 2.0 \\
Target entropy                    & 0.0 & 0.0 \\
Base (denoising steps, chunk)     & (10, 20) & (10, 20) \\
Denoising-step range              & [5, 15] & [5, 15] \\
Action-chunk range                & [10, 30] & [10, 30] \\
NFE band [lower, upper$\to$anneal]& [0.28, 0.50$\to$0.45] & [0.28, 0.50$\to$0.45] \\
NFE under / saving weight         & 0.35 / 0.08 & 0.35 / 0.08 \\
Difficulty source / weight        & \texttt{q\_std} / 0.045 & \texttt{q\_std} / 0.045 \\
Difficulty prior scale (signal)   & 0.65 (\texttt{rank\_tanh}) & 0.65 (\texttt{rank\_tanh}) \\
Grad steps / collection           & -- (offline, no rollout) & 1000 \\
\bottomrule
\end{tabular}
\end{table}

% ---------------------------------------------------------------------------
% Table 2: LIBERO-90 simulation (with DSRL baseline)
% Note: a full-width table* in a two-column layout can only float to the top of a
% page ([H] is not supported for table*). \usepackage[section]{placeins} keeps it
% within (below) this section. If your document is single-column, you may replace
% "table*" with "table" and use [H] to force it exactly below the text.
% ---------------------------------------------------------------------------
\begin{table*}[t]
\centering
\caption{Hyperparameters for the LIBERO-90 simulation tasks.
Columns t55--t59 are E-QRL (ours); the last column is the DSRL baseline.
Tasks: t55 = \emph{alphabet soup $\to$ tray}, t10 = \emph{black bowl $\to$ cabinet},
t57 = \emph{cream cheese $\to$ tray}, t59 = \emph{tomato sauce $\to$ tray}.
$^{\dagger}$Values for t10 are reconstructed from the run name (its config dict was
not saved). The DSRL baseline shares the same configuration across LIBERO tasks
(fixed steps/chunk $=10/20$); ``--'' marks E-QRL-specific entries.}
\label{tab:hp_libero}
\footnotesize
\setlength{\tabcolsep}{4pt}
\resizebox{\textwidth}{!}{%
\begin{tabular}{llllll}
\toprule
Hyperparameter & t55 & t10 & t57 & t59 & DSRL baseline \\
\midrule
Base policy                       & $\pi_0$ (libero), frozen & $\pi_0$ (libero), frozen & $\pi_0$ (libero), frozen & $\pi_0$ (libero), frozen & $\pi_0$ (libero), frozen \\
Train mode                        & Online & Online & Online & Online & Online \\
Training steps                    & $5\times10^{5}$ & $5\times10^{5}$ & $5\times10^{5}$ & $5\times10^{5}$ & $5\times10^{5}$ \\
Batch / discount $\gamma$         & 256 / 0.999 & 256 / 0.999 & 256 / 0.999 & 256 / 0.999 & 256 / 0.999 \\
LR (actor / critic / temp)        & $10^{-4}$/$3{\times}10^{-4}$/$3{\times}10^{-4}$ & $10^{-4}$/$3{\times}10^{-4}$/$3{\times}10^{-4}$ & $10^{-4}$/$3{\times}10^{-4}$/$3{\times}10^{-4}$ & $10^{-4}$/$3{\times}10^{-4}$/$3{\times}10^{-4}$ & $10^{-4}$/$3{\times}10^{-4}$/$3{\times}10^{-4}$ \\
Critic ensemble \texttt{num\_qs}  & 10 & 10 & 10 & 10 & 10 \\
Action magnitude                  & 1.0 & 1.0 & 1.0 & 1.0 & 1.0 \\
Target entropy                    & auto ($-17$) & auto ($-17$) & auto ($-17$) & auto ($-17$) & auto \\
Base (denoising steps, chunk)     & (10, 20) & (10, 20) & (10, 20) & (10, 20) & (10, 20), fixed \\
Denoising-step range              & [4, 14] & [3, 15] & [3, 15] & [3, 15] & -- \\
Action-chunk range                & [16, 42] & [16, 44] & [10, 40] & [16, 44] & -- \\
NFE band [lower, upper$\to$anneal]& [0.34, 0.50$\to$0.40] & [0.28, 0.50$\to$0.39]$^{\dagger}$ & [0.28, 0.50] & [0.28, 0.50] & -- \\
NFE under / saving weight         & 0.55 / 0.018 & 0.35 / 0.09$^{\dagger}$ & 0.35 / 0.06 & 0.35 / 0.12 & -- \\
Difficulty source / weight        & \texttt{q\_std} / 0.024 & \texttt{q\_std} / 0.06 & \texttt{q\_std} / 0.06 & \texttt{q\_std} / 0.06 & -- \\
Difficulty prior scale (signal)   & 0.82 (\texttt{rank\_tanh}) & 0.80 (\texttt{raw})$^{\dagger}$ & 0.75 (\texttt{rank\_tanh}) & 0.85 (\texttt{raw}) & -- \\
Grad steps / collection           & 20 & 20 & 20 & 20 & 20 \\
\bottomrule
\end{tabular}%
}
\end{table*}

% ---------------------------------------------------------------------------
% Table 3: ALOHA simulation (with DSRL baseline)
% ---------------------------------------------------------------------------
\begin{table}[H]
\centering
\caption{Hyperparameters for the ALOHA \texttt{transfer\_cube} simulation task.
The first column is E-QRL (ours); the second is the DSRL baseline.
$^{\ast}$ALOHA uses \texttt{difficulty\_prior\_signal\_scale}$=1.15$
(all other tasks use $1.0$); ``--'' marks E-QRL-specific entries.}
\label{tab:hp_aloha}
\small
\begin{tabular}{lll}
\toprule
Hyperparameter & \texttt{transfer\_cube} (E-QRL) & DSRL baseline \\
\midrule
Base policy                       & $\pi_0$ (ALOHA), frozen & $\pi_0$ (ALOHA), frozen \\
Train mode                        & Online & Online \\
Training steps                    & $3\times10^{6}$ & $3\times10^{6}$ \\
Batch / discount $\gamma$         & 256 / 0.999 & 256 / 0.999 \\
LR (actor / critic / temp)        & $10^{-4}$ / $3{\times}10^{-4}$ / $3{\times}10^{-4}$ & $10^{-4}$ / $3{\times}10^{-4}$ / $3{\times}10^{-4}$ \\
Critic ensemble \texttt{num\_qs}  & 10 & 10 \\
Action magnitude                  & 2.0 & 2.0 \\
Target entropy                    & 0.0 & 0.0 \\
Base (denoising steps, chunk)     & (10, 50) & (10, 50), fixed \\
Denoising-step range              & [3, 13] & -- \\
Action-chunk range                & [35, 50] & -- \\
NFE band [lower, upper$\to$anneal]& [0.08, 0.20$\to$0.17] & -- \\
NFE under / saving weight         & 0.25 / 0.09 & -- \\
Difficulty source / weight        & \texttt{q\_std} / 0.045 & -- \\
Difficulty prior scale (signal)   & 0.75 (\texttt{raw})$^{\ast}$ & -- \\
Grad steps / collection           & 20 & 20 \\
\bottomrule
\end{tabular}
\end{table}

\end{document}